\documentclass[11pt,a4paper]{article}
\usepackage[hyperref]{acl2021}
\usepackage{times}
\usepackage{latexsym}

\usepackage{microtype}
\usepackage{graphicx}
\usepackage{amsmath}
\usepackage{multirow}
\usepackage{longtable}
\usepackage{booktabs}
\usepackage{xcolor}
\usepackage{amssymb}
\usepackage{dsfont}
\usepackage{bm}

\usepackage{algorithm}
\usepackage{algorithmic}
\usepackage[switch]{lineno}

\usepackage{subcaption}
\usepackage{setspace}
\usepackage{stfloats}
\usepackage{array}

\aclfinalcopy % Uncomment this line for the final submission
\title{GoSum: Extractive Summarization of Long Documents by Reinforcement \\ Learning and Graph Organized discourse state}

\author{Junyi Bian \textsuperscript{\rm 1 \rm 8} , Xiaodi Huang \textsuperscript{\rm 2} , Hong Zhou \textsuperscript{\rm 3}, Shanfeng Zhu \textsuperscript{\rm 4 \rm 5 \rm 6 \rm 7 \rm 8}   \\
  \textsuperscript{\rm 1} School of Computer Science, Fudan University, Shanghai 200433, China\\
  \textsuperscript{\rm 2} School of Computing and Mathematics, Charles Sturt University \\ Albury, NSW 2640, Australia\\
  \textsuperscript{\rm 3} Atypon Systems, LLC, UK\\
  \textsuperscript{\rm 4} Institute of Science and Technology for Brain-Inspired Intelligence, Fudan University, China\\
  \textsuperscript{\rm 5} Key Laboratory of Computational Neuroscience and Brain-Inspired Intelligence \\ (Fudan University), Ministry of Education, Shanghai 200433, China\\
  \textsuperscript{\rm 6} MOE Frontiers Center for Brain Science, Fudan University, Shanghai 200433, China\\
  \textsuperscript{\rm 7} Zhangjiang Fudan International Innovation Center, Shanghai 200433, China\\
  \textsuperscript{\rm 8} Shanghai Key Lab of Intelligent Information Processing, \\ Fudan University, Shanghai 200433, China\\
  \texttt{\{zhusf, 20110240003\}@fudan.edu.cn, hzhou@atypon.com} \\
}
\date{}

\begin{document}
\maketitle

\begin{abstract}
Extracting summaries from long documents can be regarded as sentence classification  using the structural information of the documents.
How to use such structural information to summarize a document is challenging.
In this paper, we propose GoSum, a novel graph and reinforcement learning based extractive model for long-paper summarization.
In particular, GoSum encodes sentence states in reinforcement learning by building a heterogeneous graph for each input document at different discourse levels. An edge in the graph reflects the discourse hierarchy of a document for restraining the semantic drifts across section boundaries.
We evaluate GoSum on two datasets of scientific articles summarization: PubMed and arXiv. 
The experimental results have demonstrated that GoSum achieve state-of-the-art results compared with strong baselines of both extractive and abstractive models.
The ablation studies further validate that the performance of our GoSum benefits from the use of discourse information.

\end{abstract}

\section{Introduction}
\label{sec:intro}
% Important has been the identification of the funding and grant information

Document summarization refers to generating short and conductive summaries over given texts, which can help readers rapidly acquire essential knowledge from documents.
There are two main categories of approaches to summarization: the extractive approach and the abstractive approach.
% 抽取式模型的 summary 都来自论文原文
% 一般是通过对原文的句子进行打分筛选得到summary的。
The extractive approaches score and filter out the sentences of a given document % come from the input document,and it is usually obtained by scoring and filtering sentences of the document.%
to ensure the semantic and grammatical correctness of the selected sentences in the summary.
% 生成式模型阅读输入原文，然后理解输出得到新的文本，这类模型通常采用 seq2seq 的框架。
% 相比于抽取式模型，生成式模型的理论上限更高，也更符合人类阅读文章总结文章的行为模式。
Abstractive approaches mostly read an input text, comprehend it, and output its summary within the seq2seq framework.
This procedure is similar to humans'  summarising articles.
The theoretical upper bound on the performance of the seq2seq model is higher than what extractive approaches can achieve.
However, abstractive approaches  have the drawback of producing some meaningless and unfaithful summaries~\cite{fact_2020}. The generated summaries read smoothly with a high ROUGE score, but there is a significant gap in semantic information between them and the gold summaries.

% The challenge of long summ
In this paper, we focus on the use of extractive models for summarizing scientific literature.
Extractive summarization~\cite{matchsum_2020,zhou2018neural} has been extensively studied in short summarization datasets such as CNN/DailyMail \cite{cnndm}.
However, studies on long texts have lagged relatively behind because long document summarization is more challenging due to the following two reasons:
1) An increase in the input length expands the memory cost of the model; and
2) The complex discourse structural information about long-form documents should be taken into account.
Reading a long text, especially scientific literature, one usually glances at the discourse structure of the whole text. Once reading a section title, one roughly should know on which this section focuses. Using this structural information  of a text, one can better understand the meanings of its sentences. From the perspective of extractive summarization, it would be better to use this information for encoding sentences.
% 人类在阅读长文本尤其是科研文献的时候，通常会先了解全文的结构，当阅读某一章节的时候会知道这一章节侧重于什么内容，从而更好的理解句子的含义。从抽取式summarization的角度来考虑，在进行句子编码的时候最好也能感知这一信息。
The previous method encodes sentences and sections separately, making it difficult to capture the hierarchical structure of the document.
%之前的方法对句子和章节分别进行 encoding，很难反应论文的层次结构
In this paper, we thereby propose to use a graph neural network (GNN) to well represent the structure information of documents. The  additional benefit is that the computational complexity of GNNs is linear for long inputs.
%我们认为使用GNN能更好的进行编码，并且GNN对于长输入的计算复杂性是线性的。

% The challenge of extractive
Unlike abstractive approaches that are trained by  using available gold summaries directly, the training labels of an extractive model need to be obtained by using a search algorithm (typically greedy search) based on the gold summary provided.
This kind of label is not optimal and deterministic, i.e., the algorithm yields a single extracted label for each pair of document-abstract.
In fact, there may be many valid labels that are very similar to these suboptimal labels.
Insufficient such positive pairs may cause under-fitting~\cite{rl_2018}. 
%Further, sub-optimal labels may bring the noise into training.
These problems can be alleviated by increasing the number of samples and giving each training sample a reward from reinforcement learning (RL).
% 这个问题可以通过强化学习对更多样本采样来解决。
% and the model is trained without feedback for different training pairs.
% This brings a great challenge to extractive models.
% 不同于生成式模型直接用goldsummary训练的是，抽取式模型要间接的训练。

% Besides,
% 3) Usually, the training data of the extractive model is constructed with a search algorithm based on gold summary, which is not optimal, and the model is trained without a feedback for different training pairs.
% 1. 首先输入长度的增加，使得模型的内存捉襟见肘，
% 2. 另外 The complex discourse structural information of long-form documents needs to be taken into account
% 3. 通常，抽取式模型的训练数据是根据 gold summary 用搜索算法构建的，并不是最佳的，训练时模型没有对不同的训练pair做一个反馈。

To address the above problems, we propose a novel model called GoSum that is trained by using reinforcement learning.
Based on a given input and previously extracted sentences, GoSum generates the sentences of a summary sequentially.  
% In the process of generating a summary, we use extraction history to generate sentences sequently.
% GoSum 是一个使用强化学习训练的模型，在生成summary的过程中，我们利用抽取历史，逐一生成句子。
The process of scoring and selecting a sentence is regarded as an action in reinforcement learning.
This action is taken after the agent (the GoSum model) takes the  sentence state as input.
% 生成一个句子的过程看作是action，当前抽取完的句子加上输入的原文看作是 state。
% 对于句子和 section state 的编码, 我们考虑加入了论文的结构信息,所以我们使用了 GNN。在图的构建过程中，我们把
For encoding sentence states, we  leverage the  structure of a document.
Specifically,  we  use  a graph neural network to encode  the hierarchical structure of a document.
In more detail, we treat each sentence and section as a node of a heterogeneous graph.
 A state contains 1) a local representation of a sentence with discourse awareness, 2) the global context of a sentence within the document, and 3) information about  the extraction history.
% 对于句子和 section state 的编码我们使用了 GNN，在图的构建过程中，我们考虑加入了论文的结构信息，
% Considering RNNs-based models are usually hard to capture sentence-level long-distance dependency, especially in the case of a long document, When encoding sentences, GNN is able to use graphs to better access information about other sentences.
As such, we seamlessly integrate RL with GNN in GoSum. To summarize, our main contributions of this paper are:
1) We propose an  approach called GoSum \footnote{Source code is available  on Supplementary Files} as a novel graph-based discourse-aware extractive summarization model. GoSum can  generate a concise and informative summary operating on a subsentential discourse unit level. 2) We effectively integrate reinforcement learning with GNN under GoSum. With obtaining sufficient samples in reinforcement learning,  GoSum relies on GNN to capture discourse information about documents, particularly for the discourse hierarchy, to extract  compact summaries. 3) We have conducted comprehensive experiments to validate the performance of GoSum. GoSum has achieved state-of-the-art performance compared with strong baselines on two benchmark datasets: PubMed and arXiv.
% Moreover, we seek to observe where the performance gain of our model comes from.

% In this paper, We propose GoSum, an extractive summarization model combining reinforcement learning and graph neural networks. We achieved ideal results on both arXiv and Pubmed datasets.
% 1. 提出了 GoSum，一个结合强化学习和GNN的抽取式summ模型
% 2. 我们在 arxiv 和 pubmed 数据集上都取得了好的效果

\section{Related work}
\label{sec:relate}

% summarization 在长文本上的进展

% Gnn 模型在 summarization 上的进展
% \noindent \textbf{Long document extractive summarization.}
\subsection{Long Document Summarization}
Unlike the short-input summarization that BERT-based models~\cite{2019_bertsum} have been successfully used, studies on long document summarization struggle with long-input sequences. 
Research on abstractive models~\cite{bigbrid_2020,hepo_2021} mainly exploring different architectures of Transformer to cope with excessively long inputs.
% Although abstractive models have tried to explore different architectures of Transformer ~\cite{bigbrid_2020,hepo_2021}, 
% 生成式模型的研究侧重于改进transformer来应对过长的输入。
% Due to the unique nature of long documents such as scientific literature, they follow a standard discourse structure,
% 抽取式模型的研究则从其他角度入手。
However, the study of extractive models focus on other perspectives.
For example, long documents follow a standard discourse structure, i.e. scientific papers are written section by section to describe the background, methodology, experiment etc.
Several methods~\cite {localglobal_2019,collins2017supervised,discourse_2021} leverage such section information to guide the generation of summaries.
% LG \cite{localglobal_2019} encoding sentence by incorporating the local context with in each topic, along with the global context of the whole document.
Reinforcement learning has also successfully been applied to long document extractive summarization.
LG+RdLoss~\cite{rdloss_2020} is an improved version of LG~\cite{localglobal_2019} that constrains sentence redundancy with reinforcement learning.
Differing from LG-RdLoss, MemSum ~\cite{memsum_2022} uses extraction history~\cite{neusumm_history_2018}, and treat extractive summarization as a multi-step episodic Markov decision process.

\subsection{Graph-based Extractive Summarization}
% Early graph-based method \cite{graph0_2004} select salient sentences by a sentences similarity graph.
Early summarization solutions are graph-based unsupervised methods~\cite{graph0_2004}, relying on explicit surface features. They construct a similarity graph between sentences and formulate extractive summarization as a task of ranking nodes.
Recently, researchers use graph neural network on supervised summarization.
HSG~\cite{hsg_2020} was the first  to construct a heterogeneous graph neural network for extractive document summarization.
HahSum~\cite{hahsum_2020} considers inter-sentence redundancy in graph construction.
HEROS~\cite{discourse_2021} applies graph-based to the long text field and uses the information about input article discourse.
All these methods treat sentences and words as nodes in a graph.
Based on the RST tree, DiscoSum\cite{disco_2020} uses a graph  to capture the long-range dependencies among discourse units,  with  Elementary Discourse Units as the nodes in a graph.
 To some extent, the graph-based approach solves the quadratic computational and memory complexities encoded using Transformer and works well with the structural information of the input. Therefore, we choose to use GNNs for GoSum.
% 基于 GNN 的方法从某种程度上解决了使用 transformer 编码的 quadratic computational and memory complexities 并且能很好搭配输入的结构信息，所以我们选择使用 GNN。

\begin{figure*}[htp]
    \centering
    \includegraphics[width=2.0\columnwidth]{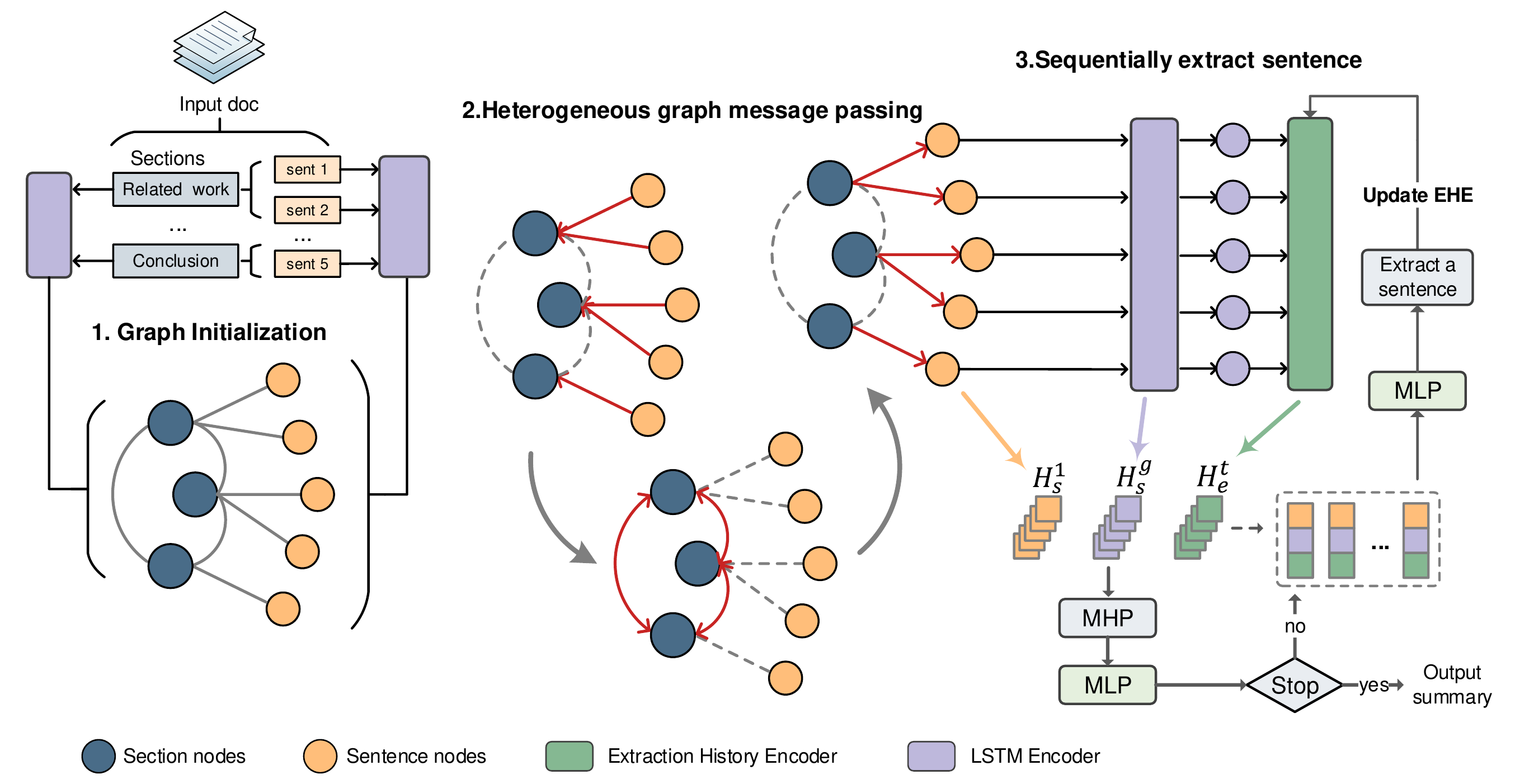}
    \caption{The overall framework of GoSum. MHP: multi-head pooling, and MLP: multi-layer perceptrons }
    \label{fig_framework}
\end{figure*}

\section{GoSum }
\label{sec:method}

Figure 1 shows the architecture of GoSum. With the input of a structural text,  GoSum starts with constructing a  graph of the text
and then generates  an embedding for the current state by using three sub-encoders: 1) The Graph-based Discourse Awareness Encoder, 2) The Global Context Encoder, and 3) The Extraction History Encoder. After this, the extractor decides whether to stop or continue the extraction based on the current embedding.

\subsection{Task Definition}
Extractive summarization is regarded as a sequence labeling task.
Denote $D=\{s_1,s_2,...,s_n\}$ as a document that consists of $n$ sentences. Extractive summarizer produces a sequence of indexes $\hat{Y} = \{\hat{y}_1,\hat{y}_2,...,\hat{y}_T\}$ to determine which sentences should be included in the summaries.
$\hat{y}_i$ denotes the index of the sentence.
Since the datasets only contain document-abstract pairs, we use beam search and automatic metric ROUGE to sample a set of oracle labels $\{Y^1,Y^2,...\}$.
Then, we keep the ROUGE score of each oracle label's corresponding summary against the abstract as a reward for reinforcement learning.

\subsection{GoSum via Policy Gradient}
From the perspective of RL for extractive summarization, we can view our GoSum model  as an agent, parameters of the network as a policy $\pi_{\theta}$, and  extracting at each step  as an action.
Given an oracle label $Y = \{y_1,y_2,...,y_T\}$, $R=(r_1,r_2,...,r_T)$ is a reward list, $r_t$ is the reward of an action to select sentence $y_t$ after the set of $\{y_1,y_2,...,y_{t-1}\}$ are already selected.
The goal of policy gradient in GoSum is to maximize objective function $\mathcal{L}(\theta)=E_{\pi_{\theta}} (R)$.
The reward value $r_t$ is the same as the ROUGE \cite{rouge_2004} score $r$ between the oracle summary and gold abstract.
\begin{align}
    r = \frac{1}{3} \left(  \text{ROUGE-1}_f + \text{ROUGE-2}_f + \text{ROUGE-L}_f  \right)
\end{align}

In reinforcement learning~\cite{rl_1992}, the policy gradient is defined as:

\begin{align}
    \nabla \mathcal{L} (\theta) = -E_{\pi_\theta} \left[  r \sum\limits_{t=1}\limits^T \nabla_{\theta} \log \pi_{\theta} (A_t|S_t,\theta)  \right]
\end{align}

where $\pi_{\theta} (A_t|S_t,\theta)$ represents the likelihood of action $A_t$ from policy net $\pi_{\theta}$ when a state is $S_t$ and the time step is $t$.
Usually, the extractive method extracts a fixed number of sentences. However, GoSum uses a stop mechanism, which  determines the point at which to stop extracting itself. So the policy likelihood can be written in the following form:
\begin{align}
    \pi(A_t|S_t,\theta) = p(\text{stop} | S_t, \theta) p(A_t|\text{stop}, S_t,\theta)
\end{align}

In each step, the policy net first outputs a probability $p_{stop}$. If $p_{stop}$ is greater than a pre-defined threshold, then the model will stop extracting, otherwise, the model continues to find the next sentence.

\subsection{State Encoder}

\subsubsection{Graph-based Discourse Awareness Encoder}

\noindent \textbf{Graph Construction:}
% 科研文献是由一个一个段落组成的，
GoSum constructs a heterogeneous graph that represents sections and sentences of a document at the discourse level.
There are only two kinds of nodes in the graph: sentence nodes and section nodes.
The way we build the graph is slightly different from the previous graph-based approach ~\cite{discourse_2021,hsg_2020,hahsum_2020} in that we discard the word nodes.
As reinforcement learning is  time-consuming, removing word nodes can significantly improve the running time  of GoSum, 
In addition, the information transferred from word nodes to sentence nodes is essentially about the representation of the sentence's local content.  Therefore, the use of a simple encoder is sufficient, such as LSTM.
We connect edges between each sentence and the section containing the sentence.
Also, a fully-connected subgraph is built among each section.

\noindent \textbf{Graph Initialization:}
After the graph is constructed, we give each node an initial representation.
Suppose that  a sentence in a document consists of $s$ words: $(sw_1,sw_2,...,sw_s)$, and the text of a section (e.g. "Related work") is composed of $c$ words: $(cw_1,cw_2,...,cw_c)$.
We   first employ  Glove\cite{glove_2014} word embeddings to embed these words, then use BiLSTM \cite{lstm_1997} with Multi-head pooling (MHP) to produce sentence representation $h_s^0$ and section representation $h_c^0$:
\begin{align}
    h_c^0 = \text{MHP} (\text{LSTM} ( \text{Glove} (cw_1,cw_2,...,cw_c) ) ) \\
    h_s^0 = \text{MHP} (\text{LSTM} ( \text{Glove} (sw_1,sw_2,...,sw_s) ) )
\end{align}

\noindent \textbf{Graph Attention Networks:}
With the available graph $G$ and its node features, we use a graph attention layer (GAT) \cite{gat_2017} to update our semantic nodes. The expressions of GAT are as follows:

\begin{align}
    e_{ij} &= \text{LeakyRELU} (W_a[W_q h_i; W_k h_j]) \\
    \alpha_{ij} &= \frac{\exp (e_{ij})}{\sum_{k\in \mathcal{N}_i} \exp (e_{ik})} \\
    h_i^{\prime} &= \sigma (\sum_{j\in \mathcal{N}_i} \alpha_{ij} W_v h_i) + h_i
\end{align}

where $W_a, W_q, W_k,$ and $ W_v$ are trainable weights, and $h_i$ is the node representation of the $i-th$ node in the graph. $\mathcal{N}_i$ is the neighbor nodes of node $i$.

\noindent \textbf{Message Passing:} We first update section nodes with their neighbor sentence nodes via the GAT and Feed Forward Net (FFN) layers:
\begin{align}
    U_{s\rightarrow c} &= \text{GAT} ({H}_c^0, {H}_s^0, {H}_s^0) \\
    {H}_c^1 &= \text{FFN}(U_{s\rightarrow c} + {H}_c^0)
\end{align}
where ${H}_s^0$ is the initialized representation of sentence nodes, and ${H}_c^0$ is for section nodes.
 $\text{GAT} ({H}_c^0, {H}_s^0, {H}_s^0)$ denotes ${H}_c^0$ as an attention query, and ${H}_s^0$ as a key and value.
% 我们继续用 section 之间的信息来更新 section 节点
We continue to update section nodes by section to section edges:

\begin{align}
    U_{c\rightarrow c} &= \text{GAT} ({H}_c^1, {H}_c^1, {H}_c^1) \\
    {H}_c^2 &= \text{FFN}(U_{c\rightarrow c} + {H}_c^1)
\end{align} \label{eq:mp3}
After a section node is updated, it already has section-level discourse information. We then pass this discourse information to each corresponding sentence node:
% 当 section节点更新完毕，section 节点已经有了 section-level discourse information，我们再把这个语义信息传递给每个对应的句子节点。
\begin{align}
    U_{c\rightarrow s} &= \text{GAT} ({H}_s^0, {H}_c^2, {H}_c^2) \\
    {H}_s^1 &= \text{FFN}(U_{c\rightarrow s} + {H}_s^0)
\end{align}
% 因为GoSum只使用了一层的 GNN，所以输出的结果是 xxx。
Since GoSum uses only one-layer GAT, the output is ${H}_s^1$.

\subsubsection{Global Context Encoder}
After that, a Bi-LSTM takes ${H}_s^1$ as input to produce sentence embeddings ${H}_s^g$ that encodes global contextual information. 
This module encodes global contextual information such as the sentence’s position in the document and information on neighboring sentences.

\subsubsection{Extraction History Encoder}
In extractive summarization, extracting sentences by an extraction history encoder is first used in NeuSum \cite{neusumm_history_2018}, in order to avoid redundancy.
% 通过观察区分已经抽取出来的句子以及还未被抽取出来的句子 EHE 会为每一个未被抽取出来的句子生成embedding，用来引导对这个句子的打分。
Comparing the extracted sentences and the remaining unextracted sentences, an extraction history encoder(EHE) generates the embedding for each of the remaining sentences. The result is used to guide the scoring of those unextracted sentences.
Our design of the extraction history encoder(EHE) in GoSum follows \cite{memsum_2022}.
It consists of a series of $N_h$ identical layers.
Each layer first performs a multi-head self-attention between the remaining sentences, followed by another multi-head self-attention performed on the sentences that have been extracted.
Two attention sublayers capture the information of both extracted and remaining sentences.
For those sentences that have not been extracted yet in time step $t$, an extraction history embedding ${H}_e^t$ is obtained.
% 对于那些尚未被抽取的句子，会得到一个抽取历史embedding。

\subsection{Extractor}
As shown in Eq(3), the extractor decides whether to stop extraction or generate the score of each remaining sentence according to the state.
% State consists of a representation of each remaining sentence is accomplished by concatenating four types of vectors including:
The state $S_t$ is described by concatenating three types of vectors:
sentence representation from discourse graph $H_s^1$,
sentence global content representation ${H}_s^g$,
and extraction history embedding ${H}_e^t$ as:
\begin{align}
    S_t = [H_s^1;H_s^g;H_e^t]
\end{align}
A multi-head pooling followed by a multi-layer perceptrons (MLP) is used to compute stop signial of extraction. Another MLP decides to extract which sentence.
% Two multi-layer perceptrons are applied to the concatenated representation. One decides whether to stop the extraction, while another decides to extract which sentence.

\begin{algorithm}[tb]
    % \setstretch{1.15}
    \caption{Training procedure in one iteration}
    \label{alg:algorithm}
    \textbf{Input}: Document-Summary pair $<D, S>$\\
    \textbf{Parameters}: Learning rate: $l$, and model parameters: $\theta$ \\
    \begin{algorithmic}[1]
        \STATE A label sequence ${Y}=\{y_1, y_2,...,y_T\}$ is sampled using beam search, with corresponding summaries having ROUGE scores $r$ against $S$.
        \STATE Obtain discourse-aware sentence embedding $H_s^1$:
        \STATE \quad Initialize Graph $H_s^0, H_c^0$
        \STATE \quad Message passing $(H_s^0, H_c^0) \rightarrow H_c^1$
        \STATE \quad Message passing $(H_c^1, H_c^1) \rightarrow H_c^2$
        \STATE \quad Message passing $(H_c^2, H_s^0) \rightarrow H_s^1$
        \STATE Obtain global-content sentence embedding $H_s^g$
        \STATE Let $t=1$.
        \WHILE{$t$ is no larger than $T$}
            \STATE Produce extraction history embedding $H_e^t$ for the remaining sentences.
            \STATE Output the probability of the sentence from the Extractor to select $y_t$ and $p_{stop}$ by using state $S_t= [ H_s^1, H_s^g, H_e^t ]$.
            \STATE Update policy gradient: $\theta \leftarrow \theta + l \cdot r \nabla \pi(A_t|S_t,\theta) $
            \STATE $t \leftarrow t + 1$
        \ENDWHILE
    \end{algorithmic}
    % \textbf{Output}: Your algorithm's output
    %
    % \begin{algorithmic}[1] %[1] enables line numbers
    %     \STATE Let $t=0$.
    %     \WHILE{condition}
    %     \STATE Do some action.
    %     \IF {conditional}
    %     \STATE Perform task A.
    %     \ELSE
    %     \STATE Perform task B.
    %     \ENDIF
    %     \ENDWHILE
    %     \STATE \textbf{return} solution
    % \end{algorithmic}
\end{algorithm}

% 模型的最后一层是两个多层感知机，其中一个决定是否停止抽取，另一个决定抽取哪个句子。

\subsection{Training}
Usually, the training samples of reinforcement learning algorithms are obtained by sampling the policy net that is currently being trained. Since the golden standard is already known at the time of training for extractive summarization, we can obtain high-quality training samples by performing beam search sampling in advance.
% 通常强化学习算法的训练样本是通过当前正在训练的 policy net 采样得到的，由于抽取式summarization在训练的时候已经知道了golden standard，我们可以通过提前进行 beam search 采样得到高质量的训练样本。
This saves the time spent on sampling and allows the model to converge more quickly.
% 这样可以节省采样花费的时间，并且让模型更快收敛。
The flow of the training process in GoSum is shown in Algorithm 1.
% 算法的流程以及细节见图1。

\section{Experiments}
\label{sec:exp}

\subsection{Summarization Datasets}
We evaluate our model on the two scientific paper datasets: PubMed and arXiv  \cite{pubmedarxiv_2018}.
Both datasets provide information about the structures of the papers.
The inputs of these datasets are the full text of scientific papers except for the abstract, and the gold summaries are the corresponding abstracts.
As can be seen from Table ~\ref{tab:dataset}, both datasets are relatively large in size, especially the arXiv dataset.
% 从表一可知，这两个数据集的规模都比较大，尤其是 arxiv 数据集。

\begin{table}[h]
  \centering
  \begin{tabular}{l|p{12mm}p{12mm}}\toprule
                                & PubMed    &   arXiv       \\ \midrule
     avg. \# sents of doc        & 89        &   207         \\
     avg. \# sents of summ       & 6.8         &   9.8          \\
     avg. \# tokens of doc       & 2730      &   5206        \\
     avg. \# tokens of summ      & 181       &   238         \\
     avg. \# sections of doc     & 6.0      &   5.5         \\\midrule
     \# Train                   & 116 937   &   202 880      \\
     \# Valid                   & 6633      &   6436        \\
     \# Test                    & 6658      &   6440        \\
     \bottomrule
\end{tabular}
\caption{The datasets we used in the expertiments.}\label{tab:dataset}
\end{table}

\subsection{Experimental Setup}

% \noindent \textbf{Evaluation Metric}
\subsubsection{Evaluation Metrics}
ROUGE score \cite{rouge_2004} is used to evaluate the model performance.
We report the F1 score of unigram, bigram overlap (ROUGE-1, ROUGE-2), and the longest common subsequence (ROUGE-L).

% \noindent \textbf{Training data Sampling}
\subsubsection{Training data Sampling}
% 我们使用 beam search 的方式进行训练数据的采集，每个 gold summary 抽取最多rouge score 最大的 15 个句子
% 对于 Pubmed 和 Arxiv 数据集我们设置最大的序列长度非别为 7 和 8.
% 抽取使用的脚本是根据 MemSum 进行修改的
The original PubMed and arXiv datasets do not provide extractive training labels.
We use beam search to obtain extractive oracle summaries. For each document-abstract pair, the algorithm generates at most 15 different summaries with the largest ROUGE score. For the PubMed and arXiv datasets, we set the maximum sequence length of extracted summaries to 7 and 8, respectively.

% \noindent \textbf{Implementation Details}
\subsubsection{Implementation Details}
Our model is trained using adam\cite{adam_2015} optimizer with the learning rate $1e-4$, $\beta_1 = 0.9$, and $\beta_2=0.999$.
GoSum and its variants are trained from 20 epochs on the both pubmed and arxiv dataset.
In each iteration, for each input document, we randomly sample one pre-prepare label for training.
Model checkpoints are saved and evaluated every 10,000 steps.
During the testing phase, the threshold of $p_{stop}$ for PubMed and arXiv is set to $0.6,$ and $ 0.45$, respectively.
GoSum and its variants are all trained on four TITAN XP GPUs.

\begin{table}[t]
  \centering
  \begin{tabular}{p{36mm}|p{8mm}p{8mm}p{8mm}}\toprule
        Models          &  R-1  &  R-2  &  R-L   \\ \midrule
        Oracle                          & 60.00 & 30.60 & 53.03  \\  \midrule
        \multicolumn{4}{c}{Extractive models}  \\ \midrule
        % &        &       &       &       &       &       &       &       &       &       \\
          Lead-10                       & 30.52 & 10.33 & 31.44  \\
          Local-Global                  & 43.77 & 17.50 & 38.71  \\
          \quad + RdLoss                & 44.01 & 17.79 & 39.09  \\
          Sent-CLF                      & 34.01 &  8.71 & 30.41  \\
          Sent-PTR                      & 42.32 & 15.63 & 38.06  \\
          HEROS                         & 47.74 & 20.46 & 42.39  \\
          \quad w/o content ranking     & 45.90 & 18.33 & 40.78  \\
          Topic-GraphSum                & 46.05 & 19.97 & 33.61  \\
          MemSum                        & 48.42 & 20.30 & 42.54  \\
          \textbf{GoSum (ours)}         & \textbf{48.61} & \textbf{20.53} & \textbf{42.80}  \\ \midrule
          \multicolumn{4}{c}{Abstractive models}  \\ \midrule
          PEGASUS                       & 44.21 & 16.95 & 38.83  \\
          BigBird-base                  & 41.22 & 16.43 & 36.96  \\
          BigBird-large                 & 46.63 & 19.02 & 41.17  \\
          Dancer                        & 45.01 & 17.60 & 40.56  \\
          HAT                           & 46.68 & 19.07 & 42.17  \\
          Hepos-Sinkhorn                & 47.87 & 20.00 & 41.50  \\
          Hepos-LSH                     & 48.24 & 20.26 & 41.78  \\
          \bottomrule
  \end{tabular}
  \caption{Results on arXiv Dataset.} \label{tab:sota_arxiv}
\end{table}

\subsubsection{Baselines}
We compare GoSum with state-of-the-art extractive methods and abstractive methods on the two datasets mentioned above.
In particular, the extractive baselines are \textit{Local-Global}~\shortcite{localglobal_2019}  that incorporates local and global contexts to extract summaries, and \textit{Local-Global+RdLoss}~\shortcite{rdloss_2020}. 
that further adds a redundancy reinforcement learning loss. 
\textit{HEROS}~\cite{discourse_2021} use heterogeneous graph-based with nodes from different discourse levels.
To ensures that the input is consistent with other baseline, we also record its results without content ranking module. 
\textit{NeuSum}~\shortcite{neusumm_history_2018} is a model that considers the extraction history. 
\textit{MemSum}~\shortcite{memsum_2022} is a reinforcement-learning-based extractive summarizer.
\textit{Sent-CLF} and \textit{Sent-PTR} \cite{clfptr_2020} are a LSTM based sentence classifier and a hierarchical seq2seq sentence pointer.

For the abstractive methods, we compare GoSum with the following methods:
\textit{PEGASUS}~\cite{pegasus_2020} is a pre-trained language model for summarization.
\textit{Dancer}~\cite{dancer_2020} is a divide-and-conquer method.
\textit{BigBird}~\shortcite{bigbrid_2020} uses sparse and windowed attentions to handle long input sequences.
\textit{Hepos}~\cite{hepo_2021} uses the efficient encoder-decoder attention with head-wise positional strides to effectively pinpoint salient information from the source.
\textit{HAT}~\cite{hat_2021} adds hierarchical attention layers to an encoder-decoder model to summarize long documents.

%.

\begin{table}[t]
  \centering
  \begin{tabular}{p{36mm}|p{8mm}p{8mm}p{8mm}}\toprule
        Models          &  R-1  &  R-2  &  R-L   \\ \midrule
        Oracle                          & 61.99 & 34.95 & 56.76  \\  \midrule
        \multicolumn{4}{c}{Extractive models}  \\ \midrule
          Lead-10                       & 37.45 & 14.19 & 34.07  \\
          Local-Global                  & 45.18 & 20.20 & 40.72  \\
          \quad + RdLoss                & 45.30 & 20.42 & 40.95  \\
          Sent-CLF                      & 45.01 & 19.91 & 41.16  \\
          Sent-PTR                      & 43.30 & 17.92 & 39.47  \\
          HEROS                         & 48.14 & 21.82 & 43.33  \\
          \quad w/o content ranking     & 46.63 & 20.63 & 42.01  \\
          Topic-GraphSum                & 48.85 & 21.76 & 35.19  \\
          NeuSum                        & 47.46 & 21.92 & 42.87  \\
          MemSum                        & 49.25 & 22.94 & 44.42  \\
          \textbf{GoSum (ours)}         & \textbf{49.83} & \textbf{23.56} & \textbf{45.10}  \\ \midrule
          \multicolumn{4}{c}{Abstractive models}  \\ \midrule
          PEGASUS                       & 45.97 & 20.15 & 41.34  \\
          BigBird-base                  & 43.70 & 19.32 & 39.99  \\
          BigBird-large                 & 46.32 & 20.65 & 42.33  \\
          Dancer                        & 46.34 & 19.97 & 42.42  \\
          HAT                           & 48.36 & 21.43 & 37.00  \\
          Hepos-Sinkhorn                & 47.93 & 20.74 & 42.58  \\
          Hepos-LSH                     & 48.12 & 21.06 & 42.72  \\ \bottomrule
  \end{tabular}
  \caption{Results on PubMed Dataset.} \label{tab:sota_pubmed}
\end{table}

\section{Results}

\subsection{Performance Comparisons}
Tables~\ref{tab:sota_arxiv} and ~\ref{tab:sota_pubmed} report the results of our model on arXiv and PubMed datasets, respectively.
On both datasets, GoSum outperforms state-of-the-art extractive and abstractive baselines. 
RL-based methods like GoSum, MemSum and LG-RdLoss show substantial performance gain, demonstrating the effectiveness of the reinforcement learning.
Compared with MemSum, GoSum has better performance. The results depend on two factors: 1) the use of the structural information from the input articles; and 2) the use of graphs to model sentences and sections. In this way, sentences can obtain more abundant information from sections, and sections can share and propagate their topical information.
GoSum has more performance improvement on PubMed dataset compared to arXiv dataset. One reason for this may be that the section information provided by the pubmed dataset is more accurate, as explained in more detail in section 5.3.
% 表1和表2分别展示了我们的模型在 pubmed 和 arxiv 数据集上的结果
% 在这两个数据集上，GoSum 取得了抽取式模型目前最好的结果，同时 GoSum 也领先于大部分的生成式模型。
% NeuSum 和 MemSum 使用了 extraction history 抽取句子，所以结果还不错
% 和 MemSum 相比，GoSum的性能更好，我们认为这取决于以下两点：1）论文的结构信息使用 2）使用 GNN 对句子和section编码，使得句子能更好的获得section间的信息。
% GoSum 在 pubmed 数据集上相对于在 arxiv 数据集上性能提升的更多。其中一个原因可能是，pubmed 数据集提供的section information 更加准确，这一点在 section 5.3 中有更详细的说明。

\subsection{Ablation Studies}

In table~\ref{tab:ablation_pubmed}, we conduct ablation studies by comparing GoSum with its variants.
%首先我们来验证图结构的有效性。

To validate the performance of the graph structure, we set the following GoSum variants:
\textbf{GoSum w/o sec2sec edges} remove section-to-section edges in graph construction, and take $H_c^1$ as a key input in Eq(13).
% GoSum w/o Graph is the model without GNN, but we still encode structural information by putting section name into an LSTM.
\textbf{GoSum w/o graph} has no graph modeling. In particular, the global contextual embedding $H_s^g$ is obtained directly using $H_s^0$. State representation $S_t$ in Eq(15) includes one more embedding $H_c^0$ to capture section information.
% we use $H_c^0$ and $H_s^0$ as section representation and sentence local content representation, respectively.
\textbf{GoSum w/o sec \& graph} does not use document structural information and graph modeling.

Improvements from \textbf{GoSum w/o graph} to \textbf{GoSum w/o sec2sec edges} demonstrate that the addition of paper structure information can slightly improve GoSum. The performance of GoSum has a greater improvement if using graphs to model the relationships between sentences and sections.

% 之后，我们来讨论不同类型的 embedding 对于模型性能的影响。
Next, we examine the effects of different embeddings on the performance of GoSum.
For \textbf{GoSum w SecE}, the extractor takes additional section representation $H_c^2$ in Eq(12).
\textbf{GoSum w/o GCE}, \textbf{GoSum w/o DLE}, and \textbf{GoSum w/o EHE}
remove Global Context Embedding $H_s^g$, Discourse aware Local sentence Embedding $H_s^1$, and Extraction History Embedding $H_s^e$ in Eq(12), respectively.

% 虽然额外增加了一个向量，但是结果反而略微下降。
Although \textbf{GoSum w SecE} adds an extra embedding, the resulting scores instead slightly decrease. This indicates that the information about section nodes has been incorporated into the local content embedding during the graph update process so that adding section embedding will be redundant with possible over-fitting.
% 当移除其他三个向量时，结果都下降了。其中移除 BBX 下降的最多， 这也说明了 discourse-aware local content sentence embedding 包含了更多有用的信息。$H_s^1$,
If the other three embeddings are removed, the performance drops. \textbf{GoSum w/o DLE} with removing $H_s^1$ results in the most decrease. This  also indicates that the discourse-aware local sentence embedding contains more useful information.
% A，B，C分别移除了，XX，YY，ZZ，结果都有下降，其中 A 下降的最明显。
% 但是结果反而有下降

\begin{table}[t]
  \centering
  \begin{tabular}{l|p{13mm}p{13mm}p{13mm}}\toprule
                              & \hfil R-1  & \hfil R-2  & \hfil R-L    \\ \midrule
        GoSum                 & \hfil 49.83 & \hfil 23.56 & \hfil 45.10   \\ \midrule
        \quad w/o sec2sec edges     & \hfil 49.72 & \hfil 23.46 & \hfil 44.99   \\
        \quad w/o graph             & \hfil 49.44 & \hfil 23.24 & \hfil 44.74   \\
        \quad w/o sec \& graph      & \hfil 49.22 & \hfil 23.02 & \hfil 44.49   \\ \midrule
        \quad w   SecE              & \hfil 49.80 & \hfil 23.56 & \hfil 45.08   \\
        \quad w/o GCE               & \hfil 48.80 & \hfil 22.34 & \hfil 44.16   \\
        \quad w/o DLE               & \hfil 48.01 & \hfil 21.84 & \hfil 43.57   \\
        \quad w/o EHE               & \hfil 49.24 & \hfil 23.09 & \hfil 44.28   \\ \bottomrule
  \end{tabular}
  \caption{Abaltion studies on PubMed dataset.} \label{tab:ablation_pubmed}
\end{table}

\subsection{What exactly enhances GoSum?}
% 在引入了强化学习和图网络，GoSum 相比于其他抽取式的方法性能提升了很多
% 这里有个疑问，使用 GCN 对于模型的提升可能源于两点，一是图的构建体现了输入的 discourse 信息，从而对模型有提升。另一点可以是图的层次结构 makes the section nodes as sink of sentence information diffusion.
% 为了解答这个疑惑，在不改变模型的情况下，我们通过逐步扰乱输入句子的 section 归属信息来观察模型的性能。section 归属信息的打乱模糊了 discourse 信息，但是图的构建依然具有层次结构。
% 从表可见，随着discourse信息的衰退，模型的性能迅速下降。
\noindent \textbf{Aspects of graph-organized discourse states:}
With the use of reinforcement learning and graph neural networks, GoSum has achieved a significant performance improvement over other extractive approaches.
A nature question of what exact reasons why GNN can enhance GoSum is raised. There may be two reasons:
One is that the use of a graph captures the input discourse information; another could be that the hierarchy of this graph makes  section nodes to be a sink of information diffusion of their sentences.
To validate these answers, we examine the  performance of GoSum by gradually scrambling the section attribution information of the input sentences without changing the other parameters of the model. The disruption of section attribution information can blur the discourse information, but the graph still keeps the hierarchical structure of documents.

\begin{figure}[t]
    \centering
    \includegraphics[width=1.1\columnwidth]{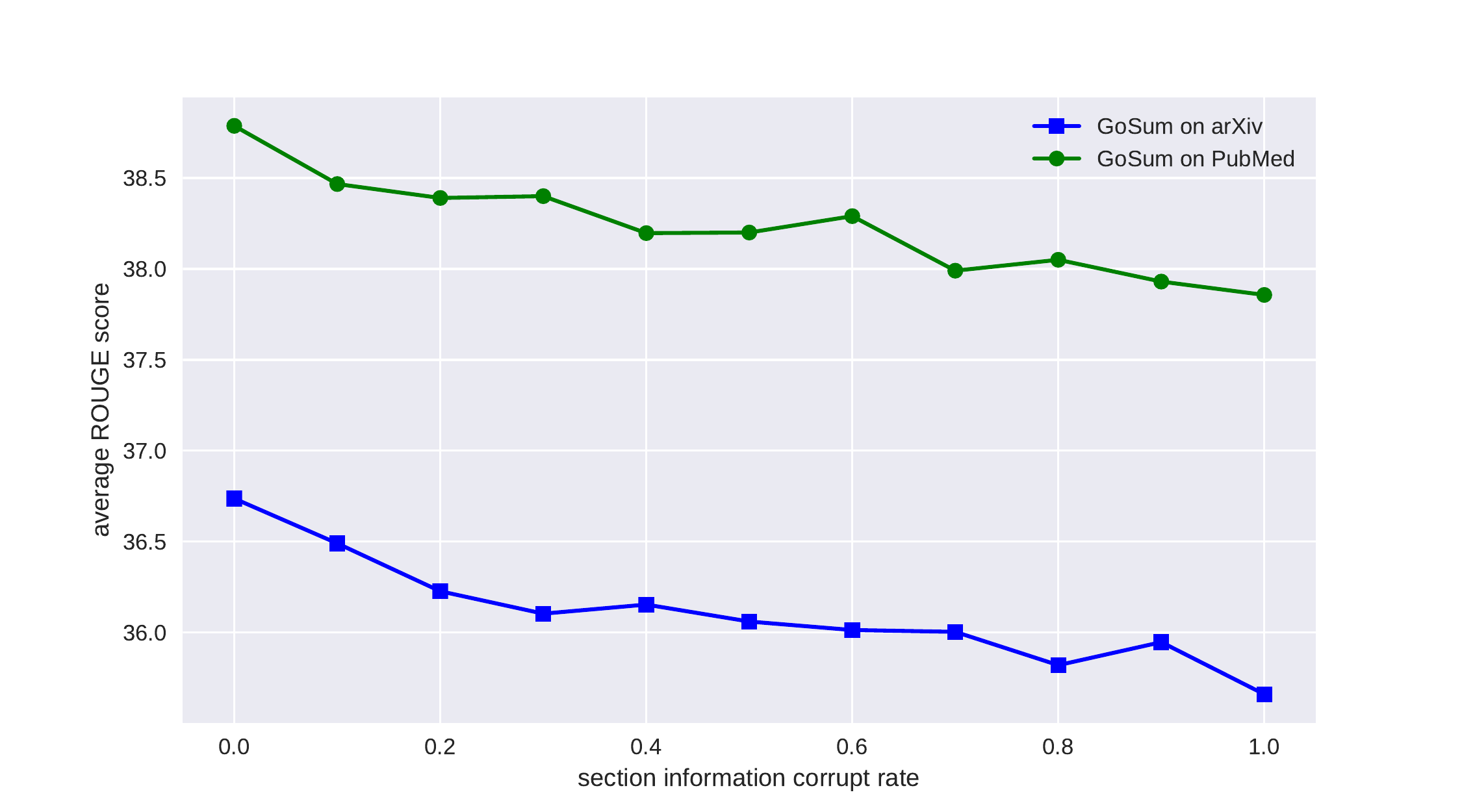} % Reduce the figure size so that it is slightly narrower than the column. Don't use precise values for figure width. This setup will avoid overfull boxes.
    \caption{GoSum performance varies as section information is corrupted at a rate (x-axis). Y-axis is the average ROUGE score. The green dots show the scores of GoSum on the PubMed dataset, while the blue dots show the results of GoSum on the arXiv dataset.}
    \label{fig:exp1}
\end{figure}

% 在这个实验中，我们按照比例打乱输入的句子的 section 归属，每次实验打乱的比例以 10% 依次递增。因为这个实验训练次数较多，所以我们对 pubmed 和 arxiv 数据集进行了采样，各自选取了 10000 个样例进行训练。
For the above reasons, we disrupt the section attribution of the input sentences proportionally, with an increment of 10\% in each experiment.
% 在完整数据集上实验太花时间了。
Since experimenting with the full data set is too time-consuming, we select 10,000 samples from each PubMed and arXiv datasets for training.
As seen from  Fig~\ref{fig:exp1}, the performance of GoSum decreases rapidly with the declining amount of discourse information.
Because of the small number of training samples and the instability of reinforcement learning, the performance of GoSum fluctuates slightly from dataset to dataset but shows a slow decreasing trend overall.
% 因为训练样本较少，并且强化学习具有不稳定性，不同数据集下的模型性能有略微的浮动，但是整体上缓慢下降。
The performance of GoSum decreases significantly at the beginning, which indicates that GoSum is sensitive to the accuracy of section information. It also confirms that accurate discourse information is required to improve the performance of GoSum.
% GoSum 的性能在刚开始下降的比较明显，这说明 GoSum 对于section信息的准确程度是比较敏感的。同时也印证了准确的discourse information 才能提升模型的性能。

\begin{table}[t]
  \centering
  \begin{tabular}{l|p{13mm}p{13mm}p{13mm}}\toprule
                    & \hfil R-1  & \hfil R-2  & \hfil R-L    \\ \midrule
                    \multicolumn{4}{c}{PubMed}     \\ \midrule
        GoSum                       & \hfil 49.83 & \hfil 23.56 & \hfil 45.10   \\
        \quad w/o SecTitle           & \hfil $\downarrow$ 0.20 & \hfil $\downarrow$ 0.11 & \hfil $\downarrow$ 0.12   \\ \midrule
                    \multicolumn{4}{c}{arXiv}     \\ \midrule
        GoSum                       & \hfil 48.61 & \hfil 20.53 & \hfil 42.80   \\
        \quad w/o SecTitle           & \hfil $\downarrow$ 0.08 & \hfil $\downarrow$ 0.03 & \hfil $\downarrow$ 0.06   \\ \bottomrule
  \end{tabular}
   \caption{Comparisions between GoSum trained with the complete datasets and those without the section titles.} \label{tab:exp2}
\end{table}

In addition to the discourse information of the literature, which divides the sentences into different sections, there are also section-specific names, such as ``introduction", ``methodology" etc.
These specific text title contexts contain semantic information, which helps to improve the performance of GoSum.
First, we set up a control model of \textbf{GoSum w/o SecTitle}, which has the same architecture as GoSum, but the section title in the training data is replaced with a meaningless text "section \#id".
The experimental results in Table~\ref{tab:exp2} show that the performance of  \textbf{GoSum w/o SecTitle} is slightly worse than that of GoSum. This indicates that the semantic information about section title text is useful but not essential. The key to performance improvement is the discourse hierarchies of documents.
Moreover, \textbf{GoSum w/o SecTitle} drops more significantly on the PubMed dataset. The difference in the performance between \textbf{GoSum w/o SecTitle}  and GoSum on the arXiv dataset is not significant, probably because the title quality of the documents in the arXiv dataset is not satisfied.

% 文献的 discourse information 除了对句子划分成不同的 section，还有 section 具体的名称，比如 “introduction”， “methodology” 等，这些具体的文本名称是含有语义信息的，从直觉上讲，也会对模型性能的提升有帮助。为此在表格中，我们又探索了 section name 对于模型性能的影响，首先，我们设置了一个对照模型 NSec，这个模型的架构和 GoSum 是一样的，但是训练数据中的 section name 用 “section #id” 来代替。实验结果显示，模型 NSec 的性能略微低于 GoSum，这说明了 section name 的语义信息虽然有用，但是并不是主要的。最重要的还是文章的结构划分。
% GoSumnsec 在 PubMed 数据集上下降的更加明显，而在 arXiv 数据集上和 GoSum 差别不大，这可能是因为 arXiv 数据集的title质量并不高。

\noindent \textbf{Aspects of reinforcement learning:}
There are two factors that can improve the performance of GoSum by using reinforcement learning: First, more sampling is performed, which is equivalent to data augmentation; and second, the model gives a feedback reward to different samples during training which helps to distinguish between good and bad samples. The experimental results on investigating the impact of these two factors on the GoSum performance are reported in Table 6. 

\textbf{w/o reward} sets rewards of all samples to 1, and the experimental results are slightly lower than those of the complete RL model.
\textbf{Complete RL} samples an average of 6.52 label sequences per document-abstract pair.
The \textbf{sample top-k} indicates that GoSum is trained with only the $k$ highest sampled label sequences of an input document.
As the number of samples increases, the  performance of GoSum improves significantly. In conclusion, the experimental results on RL verified our conjecture.

%使用了强化学习后有两点提升模型的性能，1是进行了更多的采样，相当于进行了数据增强，其次是在训练时，模型对不同样本会有一个反馈。在table 6 中的实验研究了这两点对于模型性能的影响。
%complete rl 将所有采样的样本reward都设置成了1，实验结果略低于完整的RL模型。
%完整的RL模型平均每个 docment-abstract pair 采样了 6.52 个标签序列。
%sample top-k 表示模型训练的时候只使用输入 document 的采样得分最高的 k 个标签序列。
%随着采样数量增加，模型性能提升明显。

\section{Conclusion}
\label{sec:conclude}

In this paper, we have presented a novel approach called GoSum for extracting summaries from long documents. It effectively integrates reinforcement learning with a graph neural network. In particular, we have shown how graph-organized discourse information can  be applied in reinforcement learning-based extractive summarization.
Experimental results on the arXiv and PubMed datasets have demonstrated that GoSum achieves state-of-the-art performance. The ablation experiments examine the effect of discourse information on GoSum. The results show that the performance of GoSum comes from the use of the hierarchical attribution of sentences and the semantic information about section titles of documents.
With achieving satisfactory results in scientific literature, GoSum requires  hierarchical discourse information about long texts as its inputs. In the future, we will attempt to automatically generate discourse information from documents.

\begin{table}[t]
  \centering
  \begin{tabular}{l|p{13mm}p{13mm}p{13mm}}\toprule
                    & \hfil R-1  & \hfil R-2  & \hfil R-L    \\ \midrule
        Complete RL                  & \hfil 49.83 & \hfil 23.56 & \hfil 45.10   \\
        \quad w/o reward             & \hfil 49.64 & \hfil 23.37 & \hfil 44.96   \\
        \quad sample top-1          & \hfil 49.10 & \hfil 23.00 & \hfil 44.42   \\
        \quad sample top-2          & \hfil 49.27 & \hfil 23.07 & \hfil 44.61   \\
        \quad sample top-4          & \hfil 49.64 & \hfil 23.33 & \hfil 44.96   \\ \bottomrule
  \end{tabular}
   \caption{ GoSum performance by reinforcement learning with different settings on PubMed dataset.} \label{tab:exp3}
\end{table}

% \section*{Acknowledgments}
%
% Thanks someone.

\bibliography{anthology}
\bibliographystyle{acl_natbib}

% \appendix

% \newpage

\end{document}